\documentclass[Afour,sageh,times]{sagej}

\usepackage{moreverb,url}
\usepackage[colorlinks,bookmarksopen,bookmarksnumbered,citecolor=red,urlcolor=red]{hyperref}
\newcommand\BibTeX{{\rmfamily B\kern-.05em \textsc{i\kern-.025em b}\kern-.08em
T\kern-.1667em\lower.7ex\hbox{E}\kern-.125emX}}

\usepackage{booktabs}
\usepackage{diagbox}
\usepackage{amsmath}
\usepackage{amssymb}
\usepackage{graphicx}
\usepackage{subcaption}
\usepackage{array}
\usepackage{multirow}
\usepackage{dblfloatfix}
\usepackage{enumitem}
\setlist[itemize]{leftmargin=*,label=\scalebox{.8}{\textbullet}}
\DeclareMathOperator{\argmax}{arg\,max}

\def\volumeyear{2022}
\begin{document}

\runninghead{Woo-Ri Ko et al.}

\title{Nonverbal Social Behavior Generation for Social Robots Using End-to-End Learning}

\author{Woo-Ri Ko\affilnum{1},
Minsu Jang\affilnum{1}, 
Jaeyeon Lee\affilnum{1}, and 
Jaehong Kim\affilnum{1}}
\affiliation{\affilnum{1}Electronics and Telecommunications Research Institute (ETRI), KR}

\corrauth{Woo-Ri Ko, ETRI,
218 Gajeong-ro, Yuseong-gu, Daejeon, 34129, KR.}
\email{wrko@etri.re.kr}

\begin{abstract}
To provide effective and enjoyable human-robot interaction, it is important for social robots to exhibit nonverbal behaviors, such as a \textit{handshake} or a \textit{hug}.
However, the traditional approach of reproducing pre-coded motions allows users to easily predict the reaction of the robot, giving the impression that the robot is a machine rather than a real agent.
Therefore, we propose a neural network architecture based on the Seq2Seq model that learns social behaviors from human-human interactions in an end-to-end manner.
We adopted a generative adversarial network to prevent invalid pose sequences from occurring when generating long-term behavior.
To verify the proposed method, experiments were performed using the humanoid robot Pepper in a simulated environment.
Because it is difficult to determine success or failure in social behavior generation, we propose new metrics to calculate the difference between the generated behavior and the ground-truth behavior.
We used these metrics to show how different network architectural choices affect the performance of behavior generation, and we compared the performance of learning multiple behaviors and that of learning a single behavior.
We expect that our proposed method can be used not only with home service robots, but also for guide robots, delivery robots, educational robots, and virtual robots, enabling the users to enjoy and effectively interact with the robots.
\end{abstract}

\maketitle

\section{Introduction}

To provide effective and enjoyable human-robot interactions, it is important for social robots to understand a user's behavior and generate human-like responses (\cite{wada2007living,mitsunaga2008adapting,dindo2010adaptive,salem2013err}).
For example, the robot should \textit{greet} the user when he/she comes home, \textit{high five} when the user raises his/her hand, and \textit{hug} the user when he/she is crying.
To implement these social behaviors, many studies have focused on behavior generation methods using human behavior models or predefined robot motions.
For example, \cite{breazeal1999build} developed a software architecture that exploits natural human social tendencies to enable the facial robot Kismet to engage in infant-like interactions with human caregivers.
In addition, \cite{huang2012robot} introduced a framework based on the specifications of human behavior from the social sciences to guide the generation of social behaviors for human-like robots.
Moreover, \cite{salem2012generation} proposed a control architecture that enables the humanoid robot Honda to generate gestures and synchronize speech during runtime.
Furthermore, \cite{zaraki2018development} developed an interactive sense-think-act architecture to control the humanoid robot Kaspar's behavior in a semi-autonomous manner.

However, the implementation of human behavior models or predefined robot motions requires prior knowledge of experts and human labor, which is usually costly and time-consuming.
To overcome this difficulty, recent studies have utilized data-driven learning techniques.
For example, \cite{rahmatizadeh2018vision} proposed a recurrent neural network-based architecture to learn multiple manipulation tasks based on a demonstration.
\cite{ahn2018text2action} proposed a generative model that enables robots to execute diverse actions corresponding to an input language description of human behavior.
Additionally, \cite{yoon2019robots} developed an end-to-end learning method that enables robots to learn the co-speech gestures of a humanoid robot from a TED talk.
Similarly, \cite{jonell2019learning} introduced a probabilistic generative deep learning architecture that enables robots to learn nonverbal behaviors from YouTube videos.
Furthermore, \cite{prasad2021learning} developed a framework to learn handshaking behaviors solely using data on third-person human-human interaction.

Although previous studies have produced meaningful results for the behavior generation of social robots, most studies have focused on manipulation or navigation tasks or co-speech gestures.
Several studies have been conducted on the generation of nonverbal social behaviors in robots, but they were either aimed at learning a single social behavior (\cite{prasad2021learning}), or produced invalid pose sequences when generating long-term behavior (\cite{ko2020end}).
In this study, we propose a neural network architecture based on a Seq2Seq model (\cite{sutskever2014sequence}) that can learn multiple nonverbal social behaviors.
Seq2Seq models have been predominantly used for machine translation tasks (\cite{pham2019found,cho2014learning}), and we extend their usage to generate nonverbal behavior in social robots.
In addition, we added terms in loss functions based on a generative adversarial network (GAN) (\cite{goodfellow2014generative}) to prevent invalid pose sequences from occurring when generating long-term behavior (\cite{buckchash2020variational}).

Our neural networks were trained in an end-to-end manner using the \textit{AIR-Act2Act} human-human interaction dataset introduced by \cite{ko2021air}.
The user poses were extracted from the dataset and normalized using a vector normalization method (\cite{hua2019towards}), and the robot poses were extracted and transformed into joint angles.
To validate and quantitatively evaluate the proposed method, experiments were performed using the humanoid robot Pepper in a simulated environment.
Unlike manipulation or navigation tasks, it is difficult to determine success or failure in social behavior generation, and thus we also propose new metrics that calculate the difference between the generated behavior and the ground-truth behavior.
Using these metrics, we show how different network architectural choices affect the performance of behavior generation, and we compare the performance of learning multiple behaviors and that of learning a single behavior.

\section{Problem Definition and Assumptions}

\begin{figure}[t]
    \centering
    \includegraphics[width=.95\columnwidth]{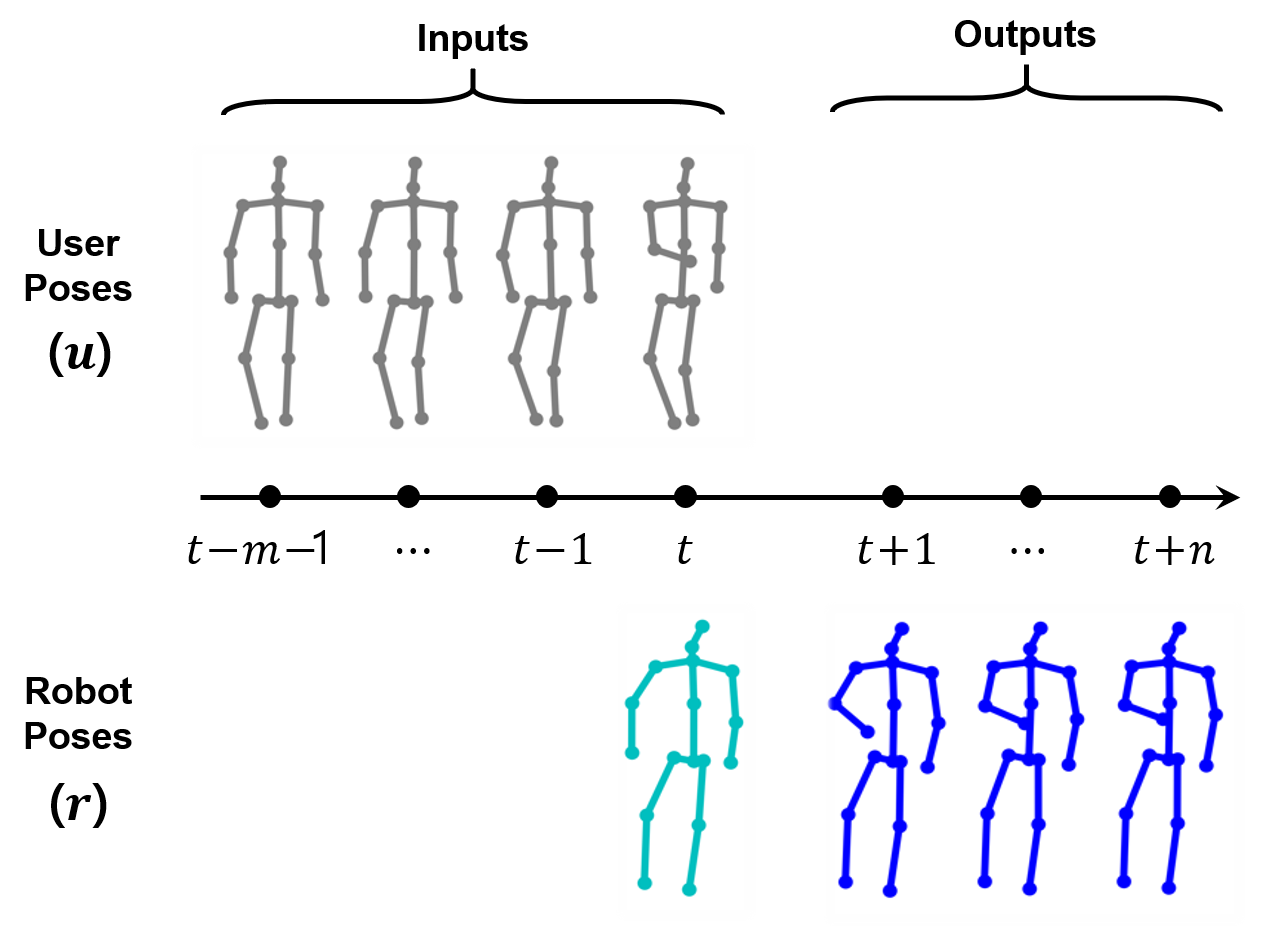}
    \caption{The generation of robot social behavior involves assigning the next robot behavior in order to respond to the current user behavior while maintaining continuity with the current robot behavior.}
    \label{fig:input-output}
\end{figure}
We consider robots that socially interact with users in home environments. 
The generation of robot social behavior involves assigning the next robot behavior $\bar{\boldsymbol{R}}_t$ to respond to the current user behavior $\boldsymbol{U}_t$ while maintaining continuity with the current robot behavior $\boldsymbol{R}_t$ at time step $t$.
Every behavior of the user and robot can be represented as a sequence of poses in the form
\begin{equation}
\begin{split}
    \text{Inputs:} \quad & \boldsymbol{U}_t = \left[\boldsymbol{u}_{t-m+1}, \boldsymbol{u}_{t-m+2}, \ldots, \boldsymbol{u}_t\right], \\
    & \boldsymbol{R}_t = \left[\boldsymbol{r}_{t}\right], \\
    \text{Outputs:} \quad & \bar{\boldsymbol{R}}_t = \left[\boldsymbol{r}_{t+1}, \boldsymbol{r}_{t+2}, \ldots, \boldsymbol{r}_{t+n}\right],
\end{split}
\end{equation}
where $\boldsymbol{u}_t$ and $\boldsymbol{r}_t$ are the poses of the user and robot at time step $t$, respectively, and $m$ and $n$ are the predefined number of poses comprising $\boldsymbol{U}_t$ and $\bar{\boldsymbol{R}}_t$, respectively.
The user pose $\boldsymbol{u}_t$ can be represented as feature points (\cite{rapantzikos2009dense}), a 2D or 3D skeleton model (\cite{redmon2017yolo9000,shotton2011real}), a depth map (\cite{wang2015action}), or an RGB image (\cite{karpathy2014large}).
To control the movement of the robot, the robot pose $\boldsymbol{r}_t$ can be represented in the same manner as a user pose, or as joint angles (\cite{yang2016repeatable}) or motor commands (\cite{levine2018learning}).

In the proposed method, we make the following assumptions:
1) The user initiates an interaction and the robot only reacts to it (the robot does not generate proactive behaviors).
2) The robot moves to a position where the user's hand is visible so that it can recognize the user's behavior, and it does not generate any behavior if the user's behavior is not recognized.
3) The robot has a 3D camera and each user pose is represented as a 3D skeleton model of the upper body. The lower body of the user is not considered because learning social behavior involves deciphering the movement of only the upper body.
4) The robot is of the humanoid type, and each robot pose is represented by the joint angles of its upper body. The lower body of the robot is not considered because of the problems associated with balancing the robot body.

\section{Proposed Method}

\begin{figure*}[t]
    \centering
    \includegraphics[width=1.\textwidth]{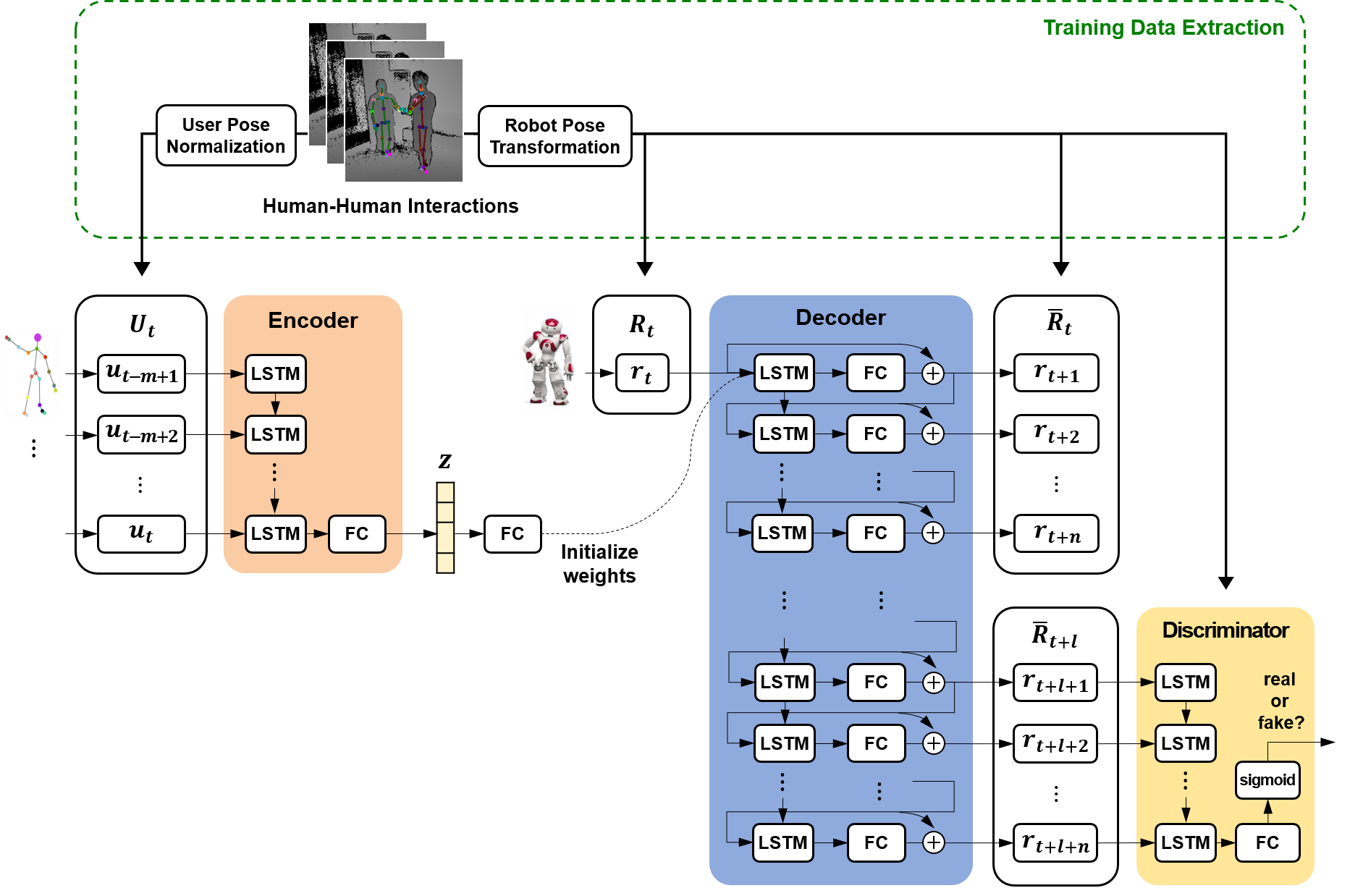}
    \caption{We propose a neural network architecture for the generation of robot social behavior consisting of an  \textit{encoder}, \textit{decoder}, and \textit{discriminator}. The \textit{encoder} encodes the current user behavior, the \textit{decoder} generates the next robot behavior according to the current user and robot behaviors, and the \textit{discriminator} aids determines if a sequence of robot poses is based on the training dataset or generated by the \textit{decoder}. The ground-truth inputs and outputs for training the \textit{encoder}, \textit{decoder}, and \textit{discriminator} are extracted from human-human interaction data.}
    \label{fig:model}
\end{figure*}

This section provides a detailed explanation of the proposed method for the generation of robot social behavior.
An overview of our method is shown in Fig. \ref{fig:model}, and it is also explained in Section \ref{sec:overview}.
The procedures for extracting training data are presented in Sections \ref{sec:extraction}, \ref{sec:userpose}, and \ref{sec:robotpose}.
Sections \ref{sec:architecture} and \ref{sec:training} describe the design, training procedure, and parameter settings of the neural network architecture.

\subsection{Overview}
\label{sec:overview}

Based on the Seq2Seq model (\cite{sutskever2014sequence}) and generative adversarial networks (GANs) (\cite{goodfellow2014generative}), we propose a neural network architecture consisting of an \textit{encoder}, \textit{decoder}, and \textit{discriminator}.
The \textit{encoder} encodes a user behavior $\boldsymbol{U}_t$ into a vector $\boldsymbol{z}$.
The \textit{decoder} generates the next robot behavior $\bar{\boldsymbol{R}}_t$ corresponding to the current robot behavior $\boldsymbol{R}_t$ and the $\boldsymbol{z}$.
For real-time interaction, the next robot behavior $\bar{\boldsymbol{R}}_t$ is generated at every $n$ time step.
The \textit{decoder} also generates the future robot behavior $\bar{\boldsymbol{R}}_{t+l}$, which is used as input for the \textit{discriminator}. 
The \textit{discriminator} determines if a sequence of robot poses is based on the training dataset or generated by the \textit{decoder}.
The ground-truth inputs and outputs for training the \textit{encoder}, \textit{decoder}, and \textit{discriminator} are extracted from human-human interaction data.

\subsection{Training Data Extraction}
\label{sec:extraction}

We first downsampled the pose data for human-human interactions to a frequency of 10 $\mathrm{Hz}$.
For instance, if the given pose data $\{t_1, t_2, t_3, \ldots\}$ are at a frequency of 30 $\mathrm{Hz}$, the downsampled pose data will be $\{t_1, t_4, t_7, \ldots\}$.
Then, the pose data of the person who initiated an interaction are normalized as a user pose $\boldsymbol{u}$, and the pose data of the person responding to the first person are transformed into a robot pose $\boldsymbol{r}$. Detailed descriptions are given in Sections \ref{sec:userpose} and \ref{sec:robotpose}.
Finally, training inputs and outputs are generated by accumulating $m$ and $n$ user and robot poses, respectively.
If the user and robot pose data are $\{\boldsymbol{u}_{t_1}, \boldsymbol{u}_{t_2},\ldots\}$ and $\{\boldsymbol{r}_{t_1}, \boldsymbol{r}_{t_2}, \ldots\}$, respectively, and $m$ and $n$ are set to 15 and 5, respectively, the first training input and output will be $\{\boldsymbol{u}_{t_1}, \boldsymbol{u}_{t_2}, \ldots, \boldsymbol{u}_{t_{15}}, \boldsymbol{r}_{t_{15}}\}$ and $\{\boldsymbol{r}_{t_{16}}, \boldsymbol{r}_{t_{17}}, \ldots, \boldsymbol{r}_{t_{20}}\}$, respectively, and the next training input and output will be $\{\boldsymbol{u}_{t_2}, \boldsymbol{u}_{t_3}, \ldots, \boldsymbol{u}_{t_{16}}, \boldsymbol{r}_{t_{16}}\}$ and $\{\boldsymbol{r}_{t_{17}}, \boldsymbol{r}_{t_{18}}, \ldots, \boldsymbol{r}_{t_{21}}\}$, respectively.

\subsection{User Pose Normalization}
\label{sec:userpose}

\begin{figure}[t]
    \centering
    \includegraphics[scale=0.5]{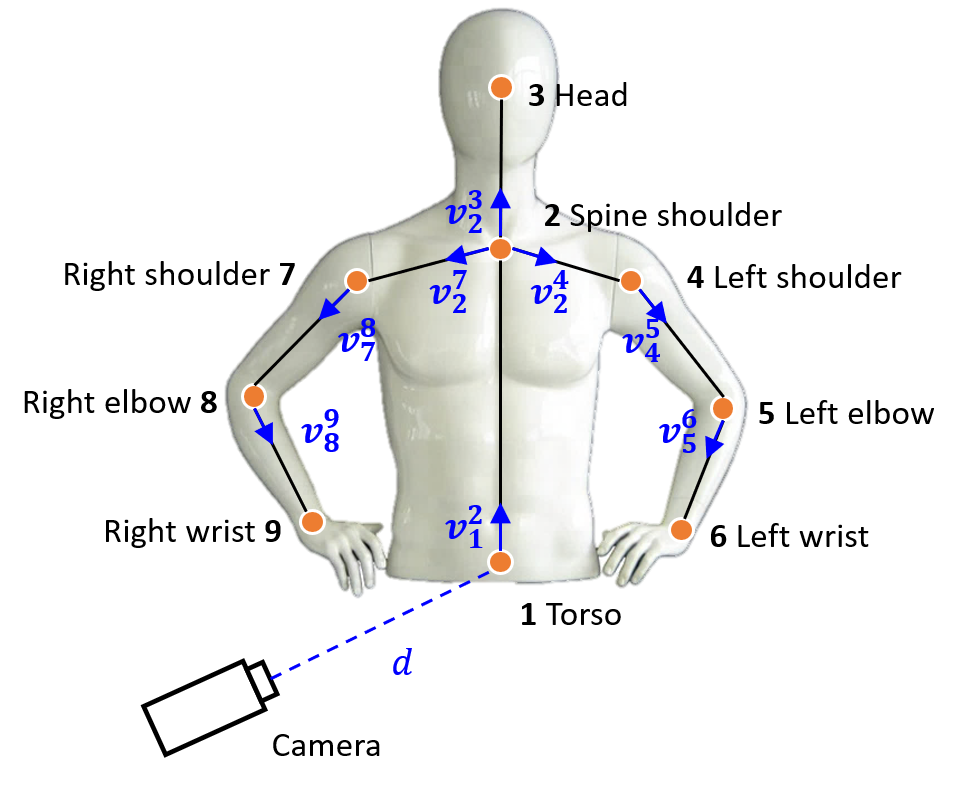}
    \caption{The nine body joints selected to represent a user pose (i.e., torso, spine shoulder, head, shoulders, elbows, and wrists) and illustration of the normalization of the user pose.}
    \label{fig:normalization}
\end{figure}

As shown in Fig. \ref{fig:normalization}, a human pose can be represented as
\begin{equation}
    \boldsymbol{P}=\left[\boldsymbol{p}_1, \boldsymbol{p}_2, \ldots, \boldsymbol{p}_9\right]^\intercal,
\label{eq:humanpose}
\end{equation}
where $\boldsymbol{p}_j=\left[x_j, y_j, z_j\right]$ indicates the 3D coordinates of the $j$-th body joint with respect to the camera.
In this study, we used nine body joints (i.e., torso, spine, shoulder, head, shoulders, elbows, and wrists) to represent a user pose, as shown in Fig. \ref{fig:normalization}.
To improve the stability and modeling performance of the neural network and emphasize active body joint movement, we adopted a vector normalization method proposed in \cite{hua2019towards}, which is expressed as
\begin{equation}
    \boldsymbol{u} = \left[\boldsymbol{v}_1^2, \boldsymbol{v}_2^3, \boldsymbol{v}_2^4, \boldsymbol{v}_4^5, \boldsymbol{v}_5^6, \boldsymbol{v}_2^7, \boldsymbol{v}_7^8, \boldsymbol{v}_8^9, d \right]^\intercal,
\label{eq:userpose}
\end{equation}
where $\boldsymbol{v}_i^j = {(\boldsymbol{p}_{j} - \boldsymbol{p}_i)} / {\lVert \boldsymbol{p}_{j} - \boldsymbol{p}_i \rVert}$ is the normalized direction vector from the $i$-th body joint to the $j$-th body joint, $d = \lVert \boldsymbol{p}_1 \rVert / d_{max}$ is the normalized distance from the camera to the torso, and $d_{max} = 5$ $\mathrm{m}$ is the maximum distance in the dataset.
Thus, the size of the normalized user pose vector is $(3 \times 8) + 1 = 25$.

\subsection{Robot Pose Transformation}
\label{sec:robotpose}

\begin{figure}[t]
    \begin{subfigure}{.49\columnwidth}
    \centering
        \includegraphics[height=5cm]{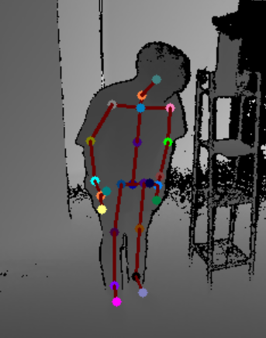}  
        \caption{\centering 3D skeleton model.}
    \end{subfigure}
    \begin{subfigure}{.49\columnwidth}
    \centering
        \includegraphics[height=5cm]{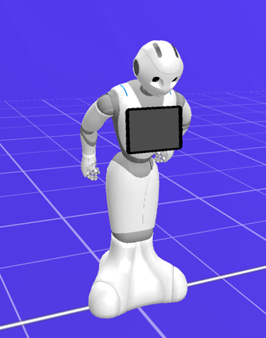}  
        \caption{\centering Pepper robot.}
    \end{subfigure}
    \caption{An example of a 3D skeleton model in the dataset (left panel) and its implementation in the Pepper robot (right panel).}
    \label{fig:robotpose}
\end{figure}
A robot pose $\boldsymbol{r}$ is represented as joint angles of the upper body.
Because we used the Pepper robot in the experiments, we analytically calculated 10 joint angles (\cite{yu2020srg}): pitches of the hip and head, pitches and rolls of the left and right shoulders, and yaws and rolls of the left and right elbows (\cite{pepper}).
As mentioned earlier, the joint angles of the lower body were not considered due to the problems associated with balancing the robot body.
Fig. \ref{fig:robotpose} shows an example of a 3D skeleton model in the dataset and its implementation in the Pepper robot.

\subsection{Neural Network Architecture}
\label{sec:architecture}

The \textit{encoder}, \textit{decoder}, and \textit{discriminator} each have a long short-term memory (LSTM) to manage time series data (\cite{hochreiter1997long}).
The \textit{encoder} takes a sequence of user poses $\boldsymbol{U}_t = \boldsymbol{u}_{(t-m+1):t}$ as input, where $m$ is set to 15 and the time interval between two adjacent user poses is 0.1 $\mathrm{s}$.
The hidden state size of its LSTM was set to 256, and the outputs of the last LSTM unit are fully connected to the layer, outputting 128 values for $\boldsymbol{z}$.
Moreover, $\boldsymbol{z}$ is fully connected to the layer producing the hidden states of the first LSTM unit of the \textit{decoder}.

The \textit{decoder} receives the current robot pose $\boldsymbol{r}_{t}$ as a seed pose and outputs a sequence of robot poses $\bar{\boldsymbol{R}}_t = \boldsymbol{r}_{(t+1):(t+n)}$, where $n$ is set to 5 in the training process.
During operation, $n$ is set to 1, meaning that behavior generation is performed at every time step.
The hidden state size of its LSTM was set to 512, and the outputs of each LSTM unit are fully connected to the layer that outputs 25 values to represent a robot pose.
We used skip connections (\cite{graves2013generating}) from the input to all LSTM units.
The \textit{decoder} also generates the future robot behavior $\bar{\boldsymbol{R}}_{t+l} = \boldsymbol{r}_{(t+l+1):(t+l+n)}$, which is used as input for the \textit{discriminator}.

The \textit{discriminator} takes a sequence of robot poses as input and outputs the probability that the sequence of robot poses is derived from the training dataset rather than the \textit{decoder}.
To help the \textit{decoder} generate competent long-term dynamics of skeleton sequences (\cite{tang2018long}), we set $l=n+25=30$ in the training process.
The hidden state size of its LSTM was set to 512, and the outputs of the last LSTM unit are fully connected to the layer, producing the probability value.

\subsection{Training}
\label{sec:training}

In our proposed neural network architecture, the combination of the \textit{encoder} and \textit{decoder} can be regarded as the generator $G$ of the GAN, where the \textit{discriminator} is the discriminator $D$ of the GAN.
To make the next behavior of the robot similar to the ground-truth behavior, we defined the loss function of $G$ as
\begin{multline}
    \mathcal{L}_{G} = \alpha_1 \cdot \mathit{MSE}\left(\langle\bar{\boldsymbol{R}}_t\rangle_{\text{gt}}, \langle\bar{\boldsymbol{R}}_t\rangle_{\text{gen}}\right) \\
    + \alpha_2 \cdot \mathit{BCE}\left(D\left(\langle\bar{\boldsymbol{R}}_{t+l}\rangle_{\text{gen}}\right), 1.0\right),
    \label{eq:lossG}
\end{multline}
where $\mathit{MSE}(x, y)$ and $\mathit{BCE}(x, y)$ are the functions used to calculate the mean square error and binary cross entropy between two vectors $x$ and $y$, respectively, $\langle\bar{\boldsymbol{R}}_t\rangle_{\text{gt}}$ and $\langle\bar{\boldsymbol{R}}_t\rangle_{\text{gen}}$ are the ground-truth and generated values of $\bar{\boldsymbol{R}}_t$, respectively, and $\alpha_1$ and $\alpha_2$ are weighting parameters that were set to 100 and 10, respectively.
In addition, to make the future robot behavior generated by $G$ appear like real robot behavior, the loss function of $D$ is defined as
\begin{multline}
    \mathcal{L}_{D} = \beta_1 \cdot \mathit{BCE}\left(\langle\bar{\boldsymbol{R}}_{t+l}\rangle_{\text{gt}}, 1.0\right) \\ 
    + \beta_2 \cdot \mathit{BCE}\left(\langle\bar{\boldsymbol{R}}_{t+l}\rangle_{\text{gen}}, 0.0\right),
    \label{eq:lossD}
\end{multline}
where $\beta_1$ and $\beta_2$ were both set to 0.5.

We iteratively trained $G$ and $D$ using the Adam optimizer (\cite{kingma2014adam}) with a mini-batch size of 100.
The learning rate was set to 0.00001, and the gradient norm was clipped to a value of 1.0 to ensure stable training.
The ``teacher forcing" technique (\cite{bengio2015scheduled}) was also adopted for fast and successful training, where the target robot pose is passed as the next input into the LSTM unit of the \textit{decoder}.
For example, we used the ground-truth output $\langle\boldsymbol{r}_{t}\rangle_{\text{gt}}$ of the $s$-th LSTM unit of the \textit{decoder} as the input to the $(s+1)$-th LSTM unit, rather than the generated output $\langle\boldsymbol{r}_{t}\rangle_{\text{gen}}$.
The probability of using $\langle\boldsymbol{r}_{t}\rangle_{\text{gt}}$ was set to 0.5.
This procedure allows the model to learn valuable information, even in the early stages of training when the quality of behavior generation is low.

\section{Experiments}
\label{sec:results}

This section presents the results of experiments performed using Pepper, a humanoid robot, in a simulated environment.
The datasets used to train and test the neural networks are described in Section \ref{sec:dataset}.
To validate the proposed behavior generation method, we describe examples of behaviors generated in seven interaction scenarios in Section \ref{sec:experiment1}.
The results of the quantitative evaluation of the proposed neural network architecture are presented in Section \ref{sec:experiment2}.

\subsection{Training and Test Datasets}
\label{sec:dataset}

\begin{table*}[t]
    \small\sf\centering
    \caption{Seven human-human interaction scenarios selected from the \textit{AIR-Act2Act} dataset (\cite{ko2021air}).}
    \begin{tabular}{c>{\raggedright\arraybackslash}p{5cm}>{\raggedright\arraybackslash}p{4.7cm}ccc}
        \toprule
        & \multicolumn{2}{c}{\textbf{Interaction Scenarios}} & \textbf{$\#$Samples} & \multicolumn{2}{c}{\textbf{Sample Lengths}} \\
        & \centering{Human1} & \centering{Human2} & ($\#$frames) & (seconds) \\
        \midrule
        1 & Enters into the service area & \textit{Bows} to Human1 & 250 & 296.6$\pm$51.3 & 9.9$\pm$1.7 \\
        2 & Walks around & \textit{Stares} at Human1 & 250 & 171.1$\pm$26.1 & 5.7$\pm$0.9 \\
        3 & Stands still without a purpose & \textit{Stares} at Human1. & 250 & 155.7$\pm$21.4 & 5.2$\pm$0.7 \\
        4 & Lifts arm to shake hands & \textit{Shakes hands} with Human1 & 250 & 149.8$\pm$18.6 & 5.0$\pm$0.6 \\
        5 & Covers face and cries & Stretches hands to \textit{hug} Human1 & 250 & 217.2$\pm$60.3 & 7.2$\pm$2.0 \\
        6 & Threatens to hit & \textit{Blocks face} with arms & 250 & 149.9$\pm$20.5 & 5.0$\pm$0.7 \\
        7 & Turns back and walks to the door & \textit{Bows} to Human1 & 250 & 172.1$\pm$35.2 & 5.7$\pm$1.2 \\
        \midrule
        \multicolumn{3}{c}{\textbf{Total}} & 1750 & 187.5$\pm$61.6 & 6.2$\pm$2.1 \\
        \bottomrule
    \end{tabular}
    \label{table:scenario}
\end{table*}

\begin{table}[t]
    \small\sf\centering
    \caption{The numbers of training and test data extracted from the selected interaction scenario data.}
    \begin{tabular}{lccc}
        \toprule
        & \textbf{Training} & \textbf{Test} & \textbf{Total} \\
        \midrule
        Interaction Samples & 1575 & 175 & 1750 \\
        Extracted Data & 116462 & 12738 & 129200 \\
        \bottomrule
    \end{tabular}
    \label{table:data}
\end{table}

To train the neural network architecture, we used the \textit{AIR-Act2Act} dataset (\cite{ko2021air}), which contains 5000 human-human interaction samples from 10 scenarios.
Considering the complexity and stability of real robot behavior implementation, seven interaction scenarios were selected, as listed in Table \ref{table:scenario}.
From the selected interaction scenarios, the social behaviors to be learned by the robots were \textit{bowing}, \textit{staring}, \textit{shaking hands}, \textit{hugging}, and \textit{blocking face}.
As summarized in Table \ref{table:scenario}, 250 samples were used to train or test each interaction scenario; the mean and standard deviation of sample lengths are 187.5 frames (6.2 seconds) and 61.6 frames (2.1 seconds), respectively.
The number of training and test data extracted by the procedure described in Section \ref{sec:extraction} are presented in Table \ref{table:data}. From 1575 interaction samples, 116462 training data were extracted, and 12738 test data were extracted from 175 interaction samples.
Each dataset containing data of a particular interaction scenario performed by a particular person exists only once in either of the training or testing datasets.
Using the training data, the neural networks were trained for 300 epochs until the performance no longer improved, which required approximately three hours with an NVIDIA RTX 3090.
At every time step in each test sample, the next robot pose was generated by feeding the algorithm a sequence of user poses and the robot pose generated in the previous time step.

\subsection{Validation of Behavior Generation}
\label{sec:experiment1}

\begin{figure*}[ht]
    \begin{subfigure}{.99\textwidth}
    \centering
        \includegraphics[width=\columnwidth]{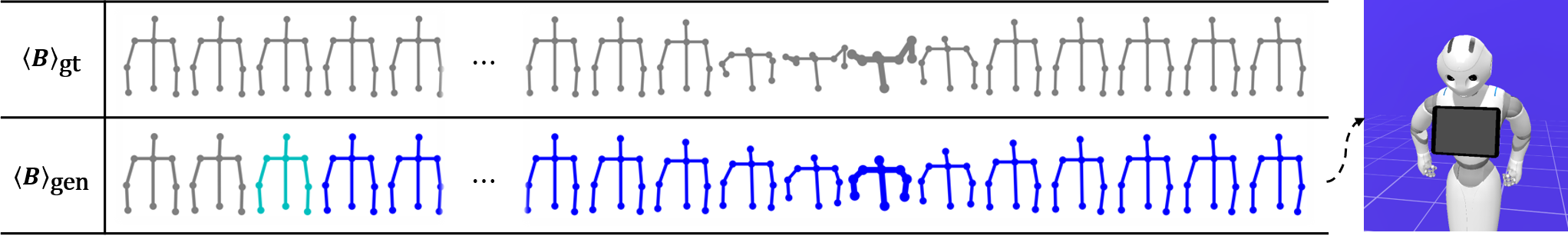}
        \caption{\centering Scenario 1 (\textit{bowing}).}
    \end{subfigure}
    \par\medskip 
    \begin{subfigure}{.99\textwidth}
    \centering
        \includegraphics[width=\columnwidth]{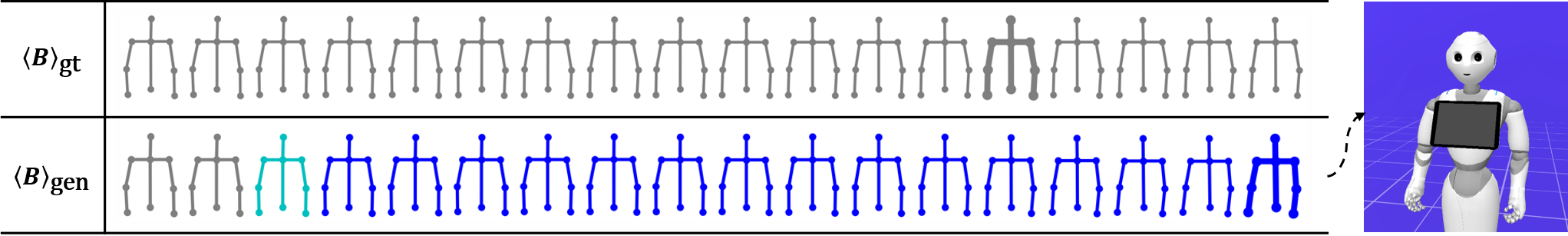} 
        \caption{\centering Scenario 2 (\textit{staring}).}
    \end{subfigure}
    \par\medskip 
    \begin{subfigure}{.99\textwidth}
    \centering
        \includegraphics[width=\columnwidth]{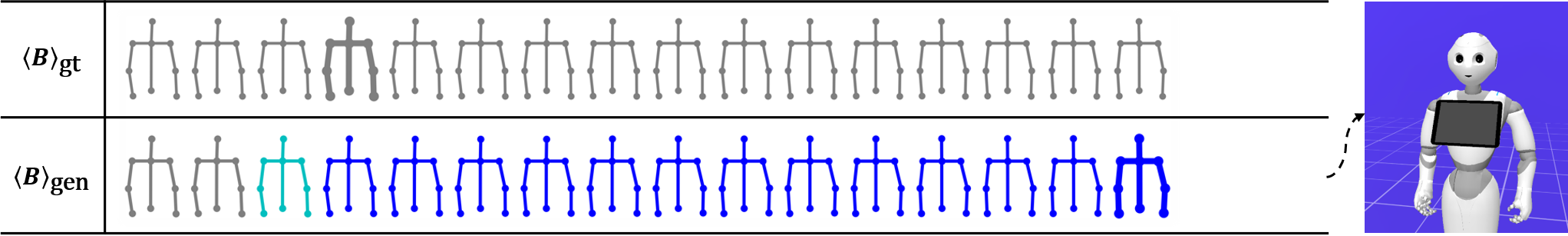} 
        \caption{\centering Scenario 3 (\textit{staring}).}
    \end{subfigure}
    \par\medskip 
    \begin{subfigure}{.99\textwidth}
    \centering
        \includegraphics[width=\columnwidth]{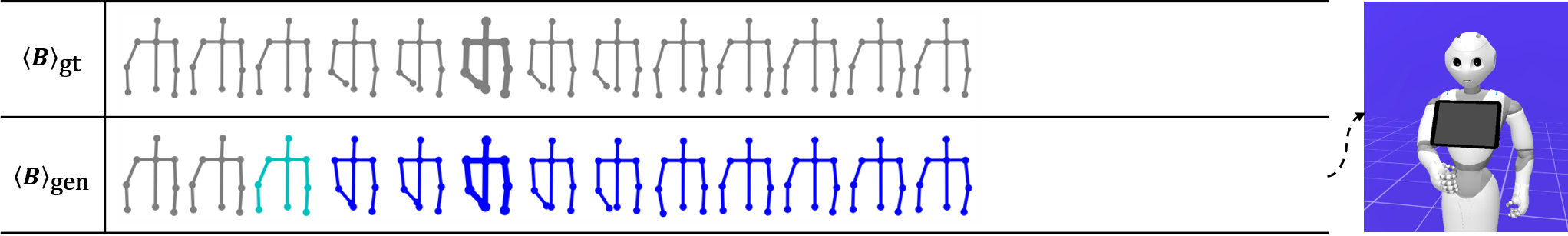} 
        \caption{\centering Scenario 4 (\textit{shaking hands}).}
    \end{subfigure}
    \par\medskip 
    \begin{subfigure}{.99\textwidth}
    \centering
        \includegraphics[width=\columnwidth]{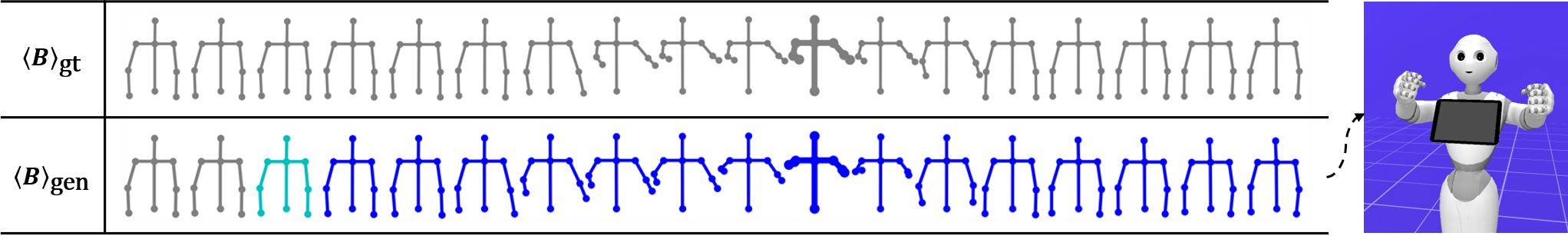}
        \caption{\centering Scenario 5 (\textit{hugging}).}
    \end{subfigure}
    \par\medskip 
    \begin{subfigure}{.99\textwidth}
    \centering
        \includegraphics[width=\columnwidth]{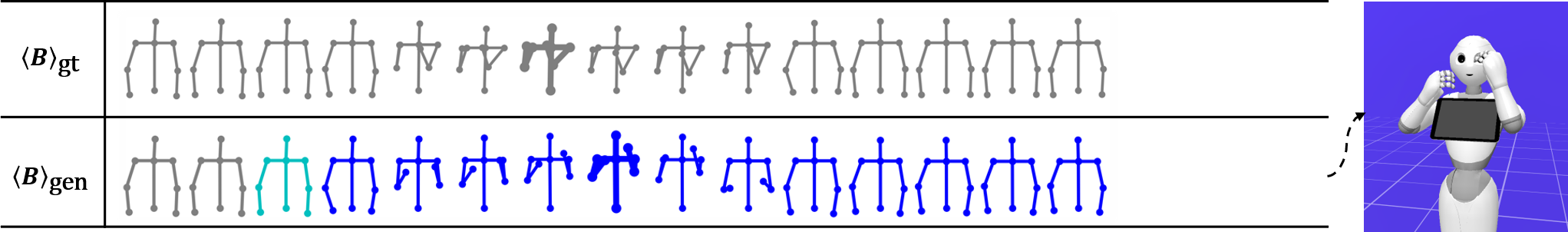} 
        \caption{\centering Scenario 6 (\textit{blocking face}).}
        \label{fig:behaviors-6}
    \end{subfigure}
    \par\medskip 
    \begin{subfigure}{.99\textwidth}
    \centering
        \includegraphics[width=\columnwidth]{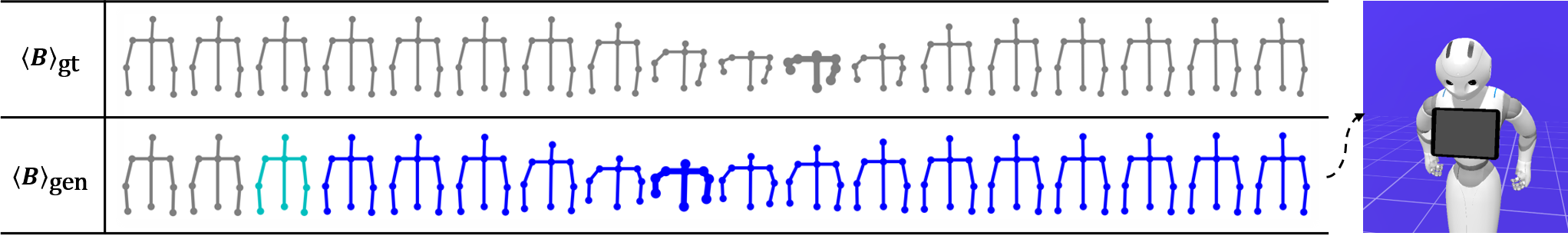} 
        \caption{\centering Scenario 7 (\textit{bowing}).}
    \end{subfigure}
    \caption{Samples of robot behaviors and key poses generated in seven interaction scenarios.}
    \label{fig:behaviors}
\end{figure*}

In this experiment, we tested robot behaviors generated in seven interaction scenarios, and Fig. \ref{fig:behaviors} shows samples of the generated robot behaviors and key poses in each scenario.
The poses in the top rows show the robot behavior $\langle\boldsymbol{B}\rangle_{\text{gt}}$ that was extracted and converted from each test interaction sample, and the bottom rows show the robot behavior $\langle\boldsymbol{B}\rangle_{\text{gen}}$ generated using the proposed method.
The gray poses indicate the ground-truth robot poses extracted and converted from the test interaction sample, and the cyan and blue poses indicate the input and output robot poses of the \textit{decoder}, respectively.
The poses in the figures were sampled such that the time interval between two adjacent poses was 0.5 $\mathrm{s}$.

Previous studies have demonstrated that key poses play an important role in behavior recognition (\cite{chaaraoui2012efficient,dhiman2020view}); we have used thick lines to indicate the key poses of $\langle\boldsymbol{B}\rangle_{\text{gt}}$ and $\langle\boldsymbol{B}\rangle_{\text{gen}}$, which are qualitatively compared.
In addition, each pose of the Pepper robot is displayed on the right with the key pose of $\langle\boldsymbol{B}\rangle_{\text{gen}}$ for each scenario.
The pose with the maximum difference from the first pose was chosen as the key pose for each behavior using
\begin{equation}
    k = \argmax_i \sum_{j} \lVert \boldsymbol{p}_{j}^i - \boldsymbol{p}_j^0 \rVert, \quad j = 3, 6, 9
\end{equation}
where $\boldsymbol{p}_{j}^i$ denotes the 3D position of the $j$-th body joint of the $i$-th robot pose with respect to the torso.
The 3rd, 6th, and 9th body joints are \textit{Head}, \textit{Left wrist}, and \textit{Right wrist}, respectively, which play more important roles in social interactions compared to other body joints.
The position of each body joint was identified by solving the kinematics equation, where the lengths between the joints were set as follows referring to \cite{pepper}.
\begin{equation}
\begin{split}
    d_1^2 &= 0.3\text{ }\mathrm{m}, \\
    d_2^3 &= d_4^5 = d_5^6 = d_7^8 = d_8^9 = 0.15\text{ }\mathrm{m}, \\
    d_2^4 &= d_2^7 = 0.08\text{ }\mathrm{m},
\end{split}
\end{equation}
where $k$ is the time index of the selected key pose and $d_i^j$ is the distance between the $i$-th and $j$-th body joints.

The experimental results showed that the behaviors generated in Scenarios 1, 4, and 5 had key poses at the same indices as the ground-truth behaviors, and the key poses were also similar.
The key poses of the behaviors generated in Scenarios 2 and 3 appeared at different indices, but the entire pose sequences were almost identical to the ground-truth behaviors.
In the behaviors generated in Scenarios 6 and 7, the key poses appeared 0.5-1 seconds late or early, but because the key poses were similar, humans may consider them almost identical to the ground-truth behaviors.

\begin{figure}[t]
    \begin{subfigure}{.54\columnwidth}
    \centering
        \includegraphics[height=3cm]{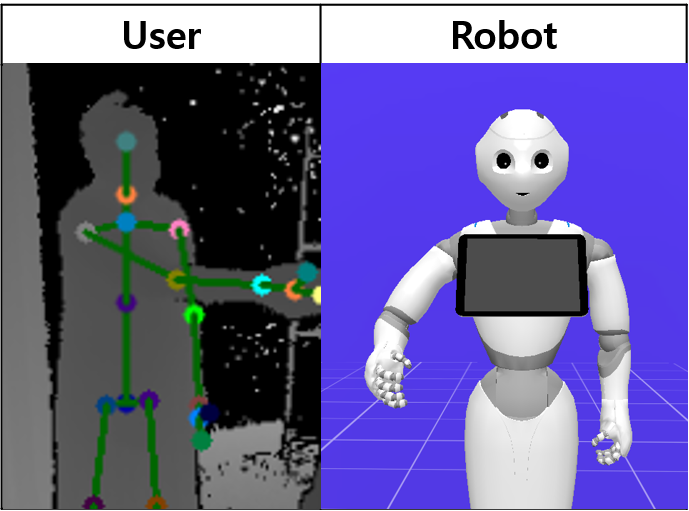}
        \caption{\centering Left position.}
    \end{subfigure}
    \begin{subfigure}{.44\columnwidth}
    \centering
        \includegraphics[height=3cm]{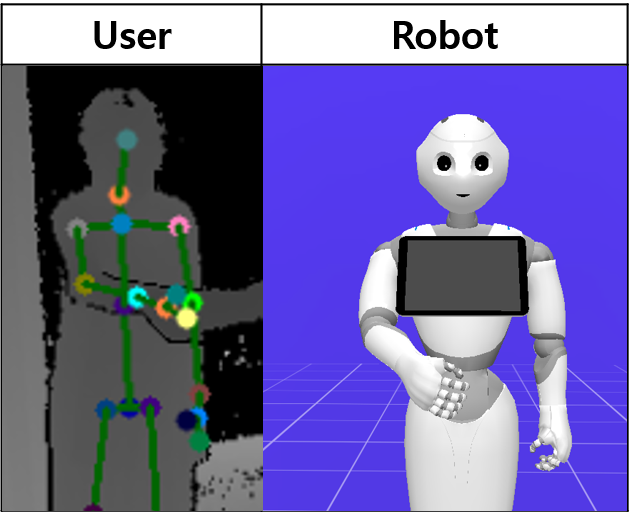}
        \caption{\centering Right position.}
    \end{subfigure}
    \par\medskip 
    \begin{subfigure}{.54\columnwidth}
    \centering
        \includegraphics[height=3cm]{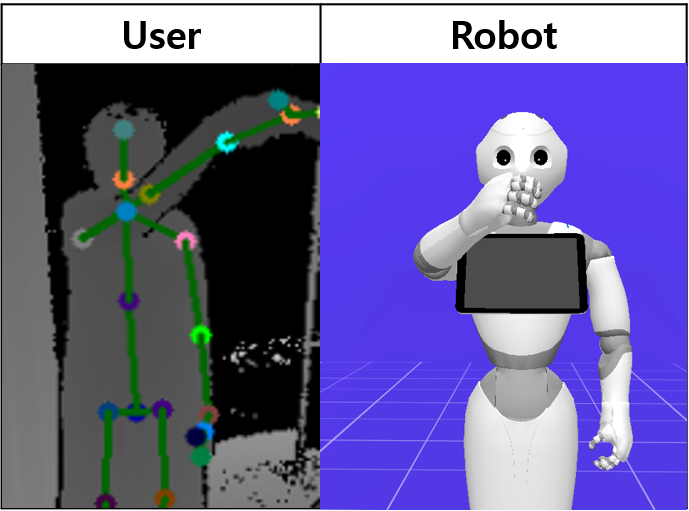}
        \caption{\centering High position.}
    \end{subfigure}
    \begin{subfigure}{.44\columnwidth}
    \centering
        \includegraphics[height=3cm]{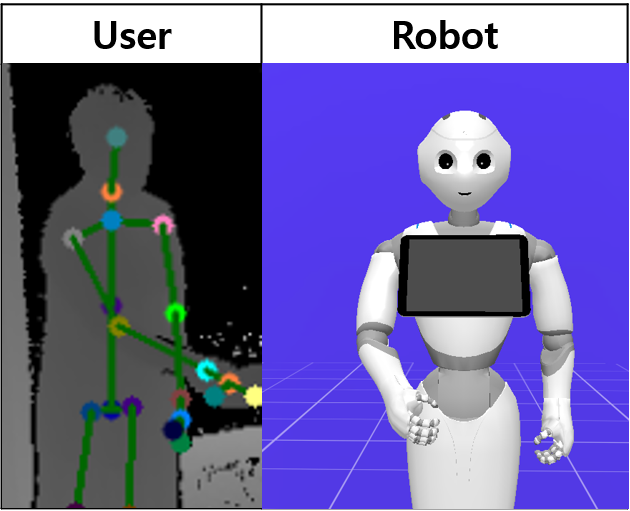} 
        \caption{\centering Low position.}
    \end{subfigure}
    \caption{Samples of robot's \textit{handshaking} behaviors generated when a user lifted his right arm to different positions.}
    \label{fig:compare}
\end{figure}

In addition, we tested robot behaviors generated when a user lifted his right arm in four different positions to shake his hands. 
Fig. \ref{fig:compare} presents samples of the generated behaviors.
We found that the robot shakes hands by moving its hand up, down, left, and right according to the position of the user's hand.
In other words, the robot can respond to the user's behavior by considering the user's posture.

\subsection{Quantitative Evaluation of Neural Network Architecture}
\label{sec:experiment2}

In this experiment, we compared the performance of the generation of robot social behavior exhibited by different neural network architectures.
Unlike in a manipulation or navigation problem, the success or failure of a social behavior task is ambiguous.
Exactly following the ground-truth pose sequence is not the only answer.
As can be seen in Fig. 5(f), the key poses of $\langle\boldsymbol{B}\rangle_{\text{gt}}$ and $\langle\boldsymbol{B}\rangle_{\text{gen}}$ may not appear in the same index. However, because the key poses are similar, most users will find that the two behaviors are nearly identical.
Therefore, when evaluating the similarity between $\langle\boldsymbol{B}\rangle_{\text{gen}}$ and $\langle\boldsymbol{B}\rangle_{\text{gt}}$, it is necessary to consider the similarity between the key poses of the two behaviors as well as the similarity between the entire pose sequences.
In particular, the movement of both hands and the degree to which the head is bowed are important in distinguishing social behaviors.
It is also important that the robot returns to its initial position when generating long-term behavior.

Therefore, we used the existing RMSE-based metric $S_1$ and our defined metrics $S_2$ and $S_3$ to determine the difference between $\langle\boldsymbol{B}\rangle_{\text{gt}}$ and $\langle\boldsymbol{B}\rangle_{\text{gen}}$ as 
\begin{equation}
    S_1 = \mathit{RMSE}\left(\langle\boldsymbol{B}\rangle_{\text{gt}}, \langle\boldsymbol{B}\rangle_{\text{gen}}\right),
\end{equation}
\begin{equation}
    S_2 = \sum_{j\in\{3,6,9\}} \lVert \langle\boldsymbol{p}_{j}^k\rangle_{\text{gt}} - \langle\boldsymbol{p}_j^k\rangle_{\text{gen}} \rVert,
\end{equation}
\begin{equation}
    S_3 = \sum_{j\in\{3,6,9\}} \lVert \langle\boldsymbol{p}_{j}^f\rangle_{\text{gt}} - \langle\boldsymbol{p}_j^f\rangle_{\text{gen}} \rVert,
\end{equation}
where $\mathit{RMSE}(A, B)$ is the root mean square error between $A$ and $B$, $\langle\boldsymbol{p}_{j}^k\rangle_{\text{gt}}$ and $\langle\boldsymbol{p}_j^k\rangle_{\text{gen}}$ are the 3D positions of the $j$-th body joint in the key poses of $\langle\boldsymbol{B}\rangle_{\text{gt}}$ and $\langle\boldsymbol{B}\rangle_{\text{gen}}$, respectively, and  $\langle\boldsymbol{p}_{j}^f\rangle_{\text{gt}}$ and $\langle\boldsymbol{p}_j^f\rangle_{\text{gen}}$ are the 3D positions of the $j$-th body joint in the final poses of $\langle\boldsymbol{B}\rangle_{\text{gt}}$ and $\langle\boldsymbol{B}\rangle_{\text{gen}}$, respectively.
The metric $S_1$ represents the sum of the errors between the poses generated at each time step.
The metric $S_2$ represents the sum of the distances between the head, left wrist, and right wrist of the two key poses, and the metric $S_3$ represents the sum of the distances between the head, left wrist, and right wrist of the two final poses.

\begin{figure}[t]
    \begin{subfigure}{.99\columnwidth}
    \centering
        \includegraphics[width=\columnwidth]{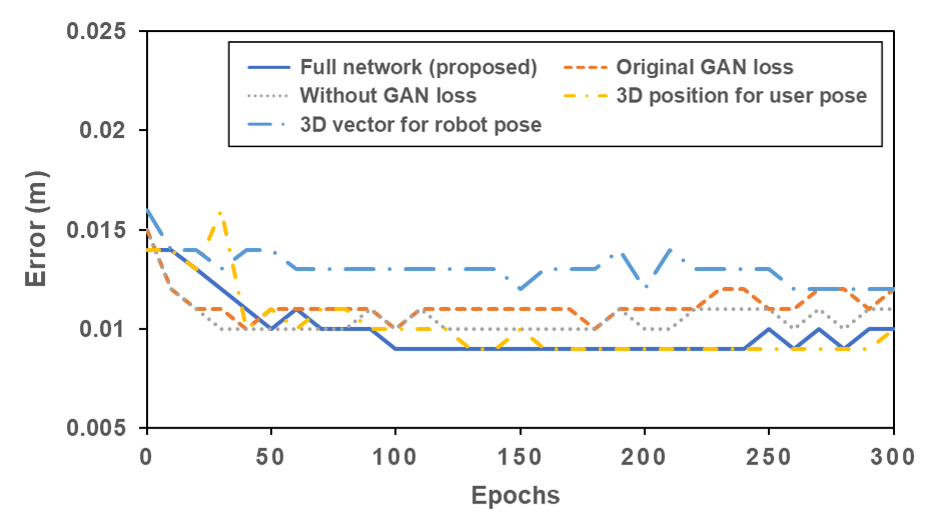}
        \caption{\centering Error in entire pose sequence ($S_1$).}
        \label{fig:performance-S1}
    \end{subfigure}
    \par\medskip 
    \begin{subfigure}{.99\columnwidth}
    \centering
        \includegraphics[width=\columnwidth]{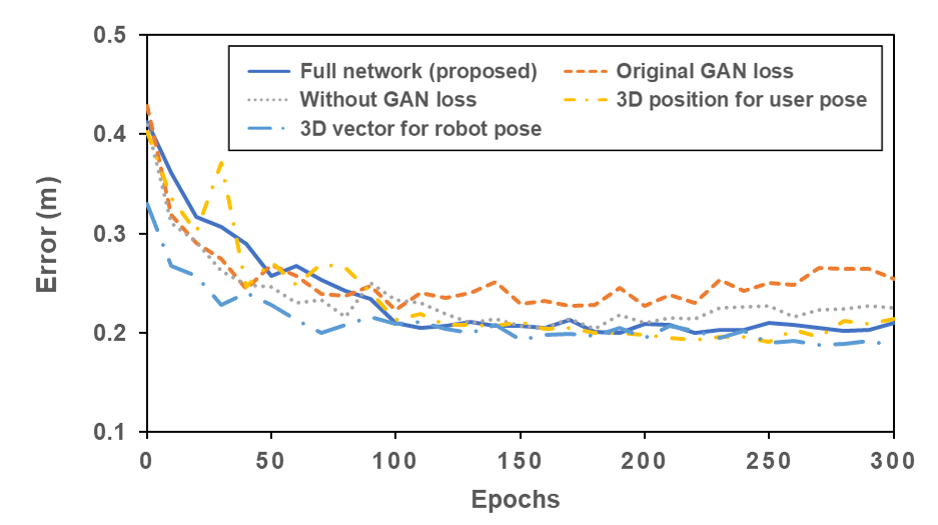}
        \caption{\centering Error in key pose ($S_2$).}
        \label{fig:performance-S2}
    \end{subfigure}
    \par\medskip 
    \begin{subfigure}{.99\columnwidth}
    \centering
        \includegraphics[width=\columnwidth]{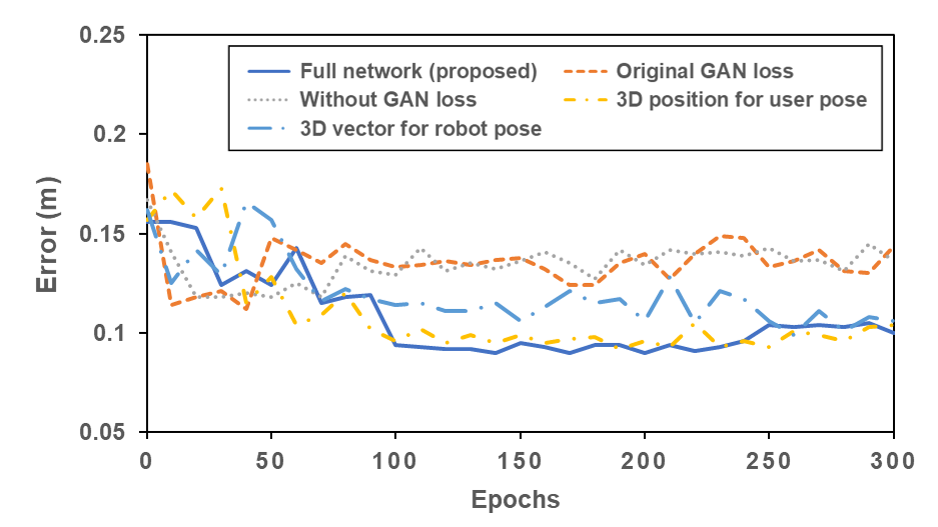}
        \caption{\centering Error in final pose ($S_3$).}
        \label{fig:performance-S3}
    \end{subfigure}
    \caption{Comparison between the performance of our neural network architecture and other architectures.}
    \label{fig:performance}
\end{figure}

We used these metrics to compare the performance of our neural network architecture with that of different architectures.
The following four architectures were prepared by removing one of the architectural components we used, and our architecture is denoted by \textit{Full network (proposed)}.
\begin{itemize}
    \item \textit{Original GAN loss} uses the next robot behavior $\bar{\boldsymbol{R}}_{t}$ instead of the future robot behavior $\bar{\boldsymbol{R}}_{t+l}$ as input to the \textit{discriminator}.
    In other words, in Equations (\ref{eq:lossG}) and (\ref{eq:lossD}), $l$ is set to 0 instead of $n+25$.
    \item \textit{Without GAN loss} omits the second term in Equation (\ref{eq:lossG}) and trains only the generator $G$, excluding the discriminator $D$.
    \item \textit{3D position for user pose} represents the user pose as the 3D positions of nine body joints (as in Equation (\ref{eq:humanpose})) instead of the direction vectors in Equation (\ref{eq:userpose}).
    \item \textit{3D vector for robot pose} represents the robot pose as the direction vectors in Equation (\ref{eq:userpose}) instead of the joint angles.
\end{itemize}

Fig. \ref{fig:performance} shows the performance of the architectures evaluated using all test data (Table \ref{table:data}).
Each architecture was trained using the same training scheme for 300 epochs.
First, the error in the entire pose sequence ($S_1$) increased in the following order: \textit{Full network (proposed)} $\approx$ \textit{3D position for the user pose} $<$ \textit{Original GAN loss} $\approx$ \textit{Without GAN loss} $<$ \textit{3D vector for robot pose}.
We found that the proposed architecture outperformed the other architectures by up to approximately 10$\%$.
The \textit{Original GAN loss} and \textit{Without GAN loss} architectures converged faster than the other architectures because the learning process is simpler.

Next, the error in the key pose ($S_2$) increased in the following order: \textit{Full network (proposed)} $\approx$ \textit{3D position for user pose} $\approx$ \textit{3D vector for robot pose} $\leq$ \textit{Without GAN loss} $<$ \textit{Original GAN loss}.
Although there was no significant difference between the performance of each  architecture, the proposed architecture outperformed the \textit{Original GAN loss} and \textit{Without GAN loss} architectures.
The representation methods of the user and robot poses did not affect the key pose errors of the generated behaviors.
However, their advantage is that they consume a small amount of computational resources because the number of features is small.

\newcolumntype{C}[1]{>{\centering\let\newline\\\arraybackslash\hspace{0pt}}m{#1}}
\begin{table*}[t]
    \small\sf\centering
    \caption{Comparison between performance of learning a single behavior and that of learning all seven behaviors.}
    \begin{tabular}{p{12em}C{1cm}C{1cm}C{1cm}C{1cm}C{1cm}C{1cm}C{1cm}C{1cm}C{1cm}}
        \toprule
        \multicolumn{1}{c}{\multirow{2}{*}{\backslashbox{\textbf{Method}}{\textbf{Error in}}}} & \textbf{Entire} & \multicolumn{4}{c}{\textbf{Key Pose}} & \multicolumn{4}{c}{\textbf{Final Pose}} \\ 
        \cmidrule(l){2-2} \cmidrule(l){3-6} \cmidrule(l){7-10}
        & $\boldsymbol{S_1}$ & Head & LWrist & RWrist & $\boldsymbol{S_2}$ & Head & LWrist & RWrist & $\boldsymbol{S_3}$, \\ \midrule 
        \multicolumn{10}{l}{\bf Learning Single Behavior} \\ \midrule
        Scenario 1 - \textit{bowing} & 0.011 & 0.078 & 0.116 & 0.081 & 0.275 & 0.092 & 0.041 & 0.046 & 0.179 \\
        Scenario 2 - \textit{staring} & 0.002 & 0.011 & 0.016 & 0.021 & 0.048 & 0.008 & 0.013 & 0.016 & 0.037 \\
        Scenario 3 - \textit{staring} & 0.000 & 0.000 & 0.001 & 0.001 & 0.002 & 0.000 & 0.001 & 0.001 & 0.002 \\
        Scenario 4 - \textit{handshaking} & 0.008 & 0.022 & 0.035 & 0.059 & 0.116 & 0.020 & 0.036 & 0.045 & 0.101 \\
        Scenario 5 - \textit{hugging} & 0.012 & 0.023 & 0.049 & 0.069 & 0.141 & 0.022 & 0.041 & 0.027 & 0.090 \\
        Scenario 6 - \textit{blocking face} & 0.011 & 0.036 & 0.089 & 0.077 & 0.202 & 0.019 & 0.041 & 0.030 & 0.090 \\
        Scenario 7 - \textit{bowing} & 0.012 & 0.088 & 0.095 & 0.074 & 0.257 & 0.060 & 0.042 & 0.036 & 0.138 \\
        \midrule
        & 0.008 & 0.037 & 0.057 & 0.055 & 0.149 & 0.032 & 0.031 & 0.029 & 0.091 \\ \midrule \midrule
        \multicolumn{10}{l}{\bf Learning All Seven Behaviors} \\ \midrule
        Scenario 1 - \textit{bowing} & 0.010 & 0.135 & 0.119 & 0.091 & 0.345 & 0.060 & 0.040 & 0.042 & 0.142 \\
        Scenario 2 - \textit{staring} & 0.004 & 0.023 & 0.035 & 0.034 & 0.092 & 0.020 & 0.031 & 0.028 & 0.080 \\
        Scenario 3 - \textit{staring} & 0.003 & 0.008 & 0.022 & 0.024 & 0.054 & 0.007 & 0.023 & 0.024 & 0.053 \\
        Scenario 4 - \textit{handshaking} & 0.008 & 0.022 & 0.058 & 0.080 & 0.160 & 0.018 & 0.032 & 0.039 & 0.089 \\
        Scenario 5 - \textit{hugging} & 0.012 & 0.020 & 0.051 & 0.067 & 0.138 & 0.021 & 0.035 & 0.029 & 0.085 \\
        Scenario 6 - \textit{blocking face} & 0.013 & 0.041 & 0.105 & 0.088 & 0.234 & 0.026 & 0.041 & 0.044 & 0.112 \\
        Scenario 7 - \textit{bowing} & 0.013 & 0.162 & 0.100 & 0.112 & 0.374 & 0.017 & 0.034 & 0.028 & 0.079 \\
        \midrule
        & 0.009 & 0.059 & 0.070 & 0.071 & 0.200 & 0.024 & 0.034 & 0.033 & 0.091 \\
        \bottomrule
    \end{tabular}
    \label{table:performance}
\end{table*}

Finally, the error in the final pose ($S_3$) increased in the following order: \textit{Full network (proposed)} $\approx$ \textit{3D position for user pose} $<$ \textit{3D vector for robot pose} $<$ \textit{Without GAN loss} $\approx$ \textit{Original GAN loss}.
The proposed architecture outperformed the other architectures by up to approximately 30$\%$ because it used GAN-based loss functions to generate competent long-term behavior.

In the next experiment, we used the full neural network architecture, but trained it on data from only one interaction scenario each time.
For each interaction scenario, the model with the smallest $S_1+S_2+S_3$ value was selected from among the models trained for 300 epochs, and the results of the performance comparison are shown in Table \ref{table:performance}.
The errors in the entire pose sequence ($S_1$) and the final pose ($S_3$) were similar when learning a single behavior and when learning all seven behaviors.
The error in the key pose ($S_2$) when learning all seven behaviors increased by approximately $30\%$ compared to when learning a single behavior. However, considering that the network learned seven times the data, it is not a large value.
Moreover, the errors in the positions of the head, left hand, and right hand were 5.9 $\mathrm{cm}$, 7.0 $\mathrm{cm}$, and 7.1 $\mathrm{cm}$, respectively, which is reasonable for social interaction (\cite{prasad2021learning}).

\section{Conclusions}
\label{sec:conclusions}
In this study, we presented an end-to-end learning-based method for learning nonverbal behaviors from human-human interactions.
A neural network architecture consisting of an \textit{encoder}, \textit{decoder}, and \textit{discriminator} was proposed.
The \textit{encoder} encoded the current user behavior, the \textit{decoder} generated the next robot behavior according to the current user and robot behaviors, and the \textit{discriminator} aided the \textit{decoder} to produce a valid pose sequence after long-term behavior was generated.
The neural networks were trained using a human-human interaction dataset \textit{AIR-Act2Act}. For this purpose, the user poses were extracted and normalized using the proposed vector normalization method, and the ground truth robot poses were extracted and transformed into joint angles of the upper body.

To validate the proposed robot behavior generation method, experiments were performed using the humanoid robot Pepper in a simulated environment.
The experimental results showed that the robot could generate five distinct social behaviors (i.e., \textit{bow}, \textit{stand}, \textit{handshake}, \textit{hug}, and \textit{block face}) and adjust their behavior according to the posture of the user.
Because it is difficult to assess success or failure in social behavior generation, we proposed two metrics to compute the similarity between the generated behavior and the ground-truth behavior.
Using these metrics, we showed that the network architectural components we used (i.e., the GAN-based loss functions, the use of future behavior as input for the \textit{discriminator}, the user pose normalization, and the robot pose transformation) improve the performance of robot behavior generation.
Moreover, the proposed method was able to learn seven social behaviors without significantly degrading the performance, which is a significant result that has not been studied so far.

With the robot generating these nonverbal social behaviors, users will feel that their behavior is understood and emotionally cared for.
Consequently, these nonverbal social behaviors can be applied not only to home service robots, but also to guide robots, delivery robots, educational robots, and virtual robots, enabling the users to enjoy and effectively interact with the robots.
However, as this study aimed to generate the behavior of a humanoid-type robot, additional research is needed to re-target the behavior so that it can be applied to other types of robots.
Additionally, as the lower body of the robot was not considered in this study because of the balancing problem, further study is needed to generate behaviors that include the lower body movements, such as moving forward or backward.
We also intend to conduct further experiments to test a robot's ability to exhibit appropriate social behaviors when deployed in the practical world and facing a human; the proposed behavior generator would be tested for its robustness to noisy input data that a robot is likely to acquire.
Moreover, by collecting and learning more interaction data, we plan to extend the number of social behaviors and complex actions that a robot can exhibit.

\begin{acks}
This work was partly supported by the Institute of Information $\&$ Communications Technology Planning $\&$ Evaluation (IITP) grant funded by the Korean government (MSIT) (No.2017-0-00162, Development of Human-care Robot Technology for Aging Society, 50$\%$) and (No.2020-0-00842, Development of Cloud Robot Intelligence for Continual Adaptation to User Reactions in Real Service Environments, 50$\%$).
\end{acks}

\bibliographystyle{SageH}
\bibliography{mybibfile}

\begin{thebibliography}{37}
\providecommand{\natexlab}[1]{#1}
\providecommand{\url}[1]{\texttt{#1}}
\providecommand{\urlprefix}{URL }
\expandafter\ifx\csname urlstyle\endcsname\relax
  \providecommand{\doi}[1]{DOI:\discretionary{}{}{}#1}\else
  \providecommand{\doi}{DOI:\discretionary{}{}{}\begingroup
  \urlstyle{rm}\Url}\fi

\bibitem[{Ahn et~al.(2018)Ahn, Ha, Choi, Yoo and Oh}]{ahn2018text2action}
Ahn H, Ha T, Choi Y, Yoo H and Oh S (2018) Text2action: Generative adversarial
  synthesis from language to action.
\newblock In: \emph{2018 IEEE International Conference on Robotics and
  Automation (ICRA)}. IEEE, pp. 5915--5920.

\bibitem[{{Aldebaran Robotics}(2020)}]{pepper}
{Aldebaran Robotics} (2020) {Pepper - Documentation}.
\newblock \url{doc.aldebaran.com/2-8/home_pepper.html}.
\newblock {[Online]}.

\bibitem[{Bengio et~al.(2015)Bengio, Vinyals, Jaitly and
  Shazeer}]{bengio2015scheduled}
Bengio S, Vinyals O, Jaitly N and Shazeer N (2015) Scheduled sampling for
  sequence prediction with recurrent neural networks.
\newblock \emph{arXiv preprint arXiv:1506.03099} .

\bibitem[{Breazeal and Scassellati(1999)}]{breazeal1999build}
Breazeal C and Scassellati B (1999) How to build robots that make friends and
  influence people.
\newblock In: \emph{Proceedings 1999 IEEE/RSJ International Conference on
  Intelligent Robots and Systems. Human and Environment Friendly Robots with
  High Intelligence and Emotional Quotients (Cat. No. 99CH36289)}, volume~2.
  IEEE, pp. 858--863.

\bibitem[{Buckchash and Raman(2020)}]{buckchash2020variational}
Buckchash H and Raman B (2020) Variational conditioning of deep recurrent
  networks for modeling complex motion dynamics.
\newblock \emph{IEEE Access} 8: 67822--67834.

\bibitem[{Chaaraoui et~al.(2012)Chaaraoui, Climent-P{\'e}rez and
  Fl{\'o}rez-Revuelta}]{chaaraoui2012efficient}
Chaaraoui AA, Climent-P{\'e}rez P and Fl{\'o}rez-Revuelta F (2012) An efficient
  approach for multi-view human action recognition based on bag-of-key-poses.
\newblock In: \emph{International Workshop on Human Behavior Understanding}.
  Springer, pp. 29--40.

\bibitem[{Cho et~al.(2014)Cho, Van~Merri{\"e}nboer, Gulcehre, Bahdanau,
  Bougares, Schwenk and Bengio}]{cho2014learning}
Cho K, Van~Merri{\"e}nboer B, Gulcehre C, Bahdanau D, Bougares F, Schwenk H and
  Bengio Y (2014) Learning phrase representations using rnn encoder-decoder for
  statistical machine translation.
\newblock \emph{arXiv preprint arXiv:1406.1078} .

\bibitem[{Dhiman and Vishwakarma(2020)}]{dhiman2020view}
Dhiman C and Vishwakarma DK (2020) View-invariant deep architecture for human
  action recognition using two-stream motion and shape temporal dynamics.
\newblock \emph{IEEE Transactions on Image Processing} 29: 3835--3844.

\bibitem[{Dindo and Schillaci(2010)}]{dindo2010adaptive}
Dindo H and Schillaci G (2010) An adaptive probabilistic approach to goal-level
  imitation learning.
\newblock In: \emph{2010 IEEE/RSJ International Conference on Intelligent
  Robots and Systems}. IEEE, pp. 4452--4457.

\bibitem[{Goodfellow et~al.(2014)Goodfellow, Pouget-Abadie, Mirza, Xu,
  Warde-Farley, Ozair, Courville and Bengio}]{goodfellow2014generative}
Goodfellow I, Pouget-Abadie J, Mirza M, Xu B, Warde-Farley D, Ozair S,
  Courville A and Bengio Y (2014) Generative adversarial nets.
\newblock \emph{Advances in neural information processing systems} 27.

\bibitem[{Graves(2013)}]{graves2013generating}
Graves A (2013) Generating sequences with recurrent neural networks.
\newblock \emph{arXiv preprint arXiv:1308.0850} .

\bibitem[{Hochreiter and Schmidhuber(1997)}]{hochreiter1997long}
Hochreiter S and Schmidhuber J (1997) Long short-term memory.
\newblock \emph{Neural computation} 9(8): 1735--1780.

\bibitem[{Hua et~al.(2019)Hua, Shi, Nan, Wang, Chen and Lian}]{hua2019towards}
Hua M, Shi F, Nan Y, Wang K, Chen H and Lian S (2019) Towards more realistic
  human-robot conversation: A seq2seq-based body gesture interaction system.
\newblock In: \emph{2019 IEEE/RSJ International Conference on Intelligent
  Robots and Systems (IROS)}. IEEE, pp. 1393--1400.

\bibitem[{Huang and Mutlu(2012)}]{huang2012robot}
Huang CM and Mutlu B (2012) Robot behavior toolkit: generating effective social
  behaviors for robots.
\newblock In: \emph{2012 7th ACM/IEEE International Conference on Human-Robot
  Interaction (HRI)}. IEEE, pp. 25--32.

\bibitem[{Jonell et~al.(2019)Jonell, Kucherenko, Ekstedt and
  Beskow}]{jonell2019learning}
Jonell P, Kucherenko T, Ekstedt E and Beskow J (2019) Learning non-verbal
  behavior for a social robot from youtube videos.
\newblock In: \emph{ICDL-EpiRob Workshop on Naturalistic Non-Verbal and
  Affective Human-Robot Interactions, Oslo, Norway, August 19, 2019}.

\bibitem[{Karpathy et~al.(2014)Karpathy, Toderici, Shetty, Leung, Sukthankar
  and Fei-Fei}]{karpathy2014large}
Karpathy A, Toderici G, Shetty S, Leung T, Sukthankar R and Fei-Fei L (2014)
  Large-scale video classification with convolutional neural networks.
\newblock In: \emph{Proceedings of the IEEE conference on Computer Vision and
  Pattern Recognition}. pp. 1725--1732.

\bibitem[{Kingma and Ba(2014)}]{kingma2014adam}
Kingma DP and Ba J (2014) Adam: A method for stochastic optimization.
\newblock \emph{arXiv preprint arXiv:1412.6980} .

\bibitem[{Ko et~al.(2021)Ko, Jang, Lee and Kim}]{ko2021air}
Ko WR, Jang M, Lee J and Kim J (2021) Air-act2act: Human--human interaction
  dataset for teaching non-verbal social behaviors to robots.
\newblock \emph{The International Journal of Robotics Research} 40(4-5):
  691--697.

\bibitem[{Ko et~al.(2020)Ko, Lee, Jang and Kim}]{ko2020end}
Ko WR, Lee J, Jang M and Kim J (2020) End-to-end learning of social behaviors
  for humanoid robots.
\newblock In: \emph{2020 IEEE International Conference on Systems, Man, and
  Cybernetics (SMC)}. IEEE, pp. 1200--1205.

\bibitem[{Levine et~al.(2018)Levine, Pastor, Krizhevsky, Ibarz and
  Quillen}]{levine2018learning}
Levine S, Pastor P, Krizhevsky A, Ibarz J and Quillen D (2018) Learning
  hand-eye coordination for robotic grasping with deep learning and large-scale
  data collection.
\newblock \emph{The International Journal of Robotics Research} 37(4-5):
  421--436.

\bibitem[{Mitsunaga et~al.(2008)Mitsunaga, Smith, Kanda, Ishiguro and
  Hagita}]{mitsunaga2008adapting}
Mitsunaga N, Smith C, Kanda T, Ishiguro H and Hagita N (2008) Adapting robot
  behavior for human--robot interaction.
\newblock \emph{IEEE Transactions on Robotics} 24(4): 911--916.

\bibitem[{Pham et~al.(2019)Pham, Liang, Manzini, Morency and
  P{\'o}czos}]{pham2019found}
Pham H, Liang PP, Manzini T, Morency LP and P{\'o}czos B (2019) Found in
  translation: Learning robust joint representations by cyclic translations
  between modalities.
\newblock In: \emph{Proceedings of the AAAI Conference on Artificial
  Intelligence}, volume~33. pp. 6892--6899.

\bibitem[{Prasad et~al.(2021)Prasad, Stock-Homburg and
  Peters}]{prasad2021learning}
Prasad V, Stock-Homburg R and Peters J (2021) Learning human-like hand reaching
  for human-robot handshaking.
\newblock \emph{arXiv preprint arXiv:2103.00616} .

\bibitem[{Rahmatizadeh et~al.(2018)Rahmatizadeh, Abolghasemi, B{\"o}l{\"o}ni
  and Levine}]{rahmatizadeh2018vision}
Rahmatizadeh R, Abolghasemi P, B{\"o}l{\"o}ni L and Levine S (2018)
  Vision-based multi-task manipulation for inexpensive robots using end-to-end
  learning from demonstration.
\newblock In: \emph{2018 IEEE international conference on robotics and
  automation (ICRA)}. IEEE, pp. 3758--3765.

\bibitem[{Rapantzikos et~al.(2009)Rapantzikos, Avrithis and
  Kollias}]{rapantzikos2009dense}
Rapantzikos K, Avrithis Y and Kollias S (2009) Dense saliency-based
  spatiotemporal feature points for action recognition.
\newblock In: \emph{2009 IEEE Conference on Computer Vision and Pattern
  Recognition}. IEEE, pp. 1454--1461.

\bibitem[{Redmon and Farhadi(2017)}]{redmon2017yolo9000}
Redmon J and Farhadi A (2017) Yolo9000: better, faster, stronger.
\newblock In: \emph{Proceedings of the IEEE conference on computer vision and
  pattern recognition}. pp. 7263--7271.

\bibitem[{Salem et~al.(2013)Salem, Eyssel, Rohlfing, Kopp and
  Joublin}]{salem2013err}
Salem M, Eyssel F, Rohlfing K, Kopp S and Joublin F (2013) To err is human
  (-like): Effects of robot gesture on perceived anthropomorphism and
  likability.
\newblock \emph{International Journal of Social Robotics} 5(3): 313--323.

\bibitem[{Salem et~al.(2012)Salem, Kopp, Wachsmuth, Rohlfing and
  Joublin}]{salem2012generation}
Salem M, Kopp S, Wachsmuth I, Rohlfing K and Joublin F (2012) Generation and
  evaluation of communicative robot gesture.
\newblock \emph{International Journal of Social Robotics} 4(2): 201--217.

\bibitem[{Shotton et~al.(2011)Shotton, Fitzgibbon, Cook, Sharp, Finocchio,
  Moore, Kipman and Blake}]{shotton2011real}
Shotton J, Fitzgibbon A, Cook M, Sharp T, Finocchio M, Moore R, Kipman A and
  Blake A (2011) Real-time human pose recognition in parts from single depth
  images.
\newblock In: \emph{CVPR 2011}. IEEE, pp. 1297--1304.

\bibitem[{Sutskever et~al.(2014)Sutskever, Vinyals and
  Le}]{sutskever2014sequence}
Sutskever I, Vinyals O and Le QV (2014) Sequence to sequence learning with
  neural networks.
\newblock In: \emph{Advances in neural information processing systems}. pp.
  3104--3112.

\bibitem[{Tang et~al.(2018)Tang, Ma, Liu and Zheng}]{tang2018long}
Tang Y, Ma L, Liu W and Zheng W (2018) Long-term human motion prediction by
  modeling motion context and enhancing motion dynamic.
\newblock \emph{arXiv preprint arXiv:1805.02513} .

\bibitem[{Wada and Shibata(2007)}]{wada2007living}
Wada K and Shibata T (2007) Living with seal robots—its sociopsychological
  and physiological influences on the elderly at a care house.
\newblock \emph{IEEE transactions on robotics} 23(5): 972--980.

\bibitem[{Wang et~al.(2015)Wang, Li, Gao, Zhang, Tang and
  Ogunbona}]{wang2015action}
Wang P, Li W, Gao Z, Zhang J, Tang C and Ogunbona PO (2015) Action recognition
  from depth maps using deep convolutional neural networks.
\newblock \emph{IEEE Transactions on Human-Machine Systems} 46(4): 498--509.

\bibitem[{Yang et~al.(2016)Yang, Sasaki, Suzuki, Kase, Sugano and
  Ogata}]{yang2016repeatable}
Yang PC, Sasaki K, Suzuki K, Kase K, Sugano S and Ogata T (2016) Repeatable
  folding task by humanoid robot worker using deep learning.
\newblock \emph{IEEE Robotics and Automation Letters} 2(2): 397--403.

\bibitem[{Yoon et~al.(2019)Yoon, Ko, Jang, Lee, Kim and Lee}]{yoon2019robots}
Yoon Y, Ko WR, Jang M, Lee J, Kim J and Lee G (2019) Robots learn social
  skills: End-to-end learning of co-speech gesture generation for humanoid
  robots.
\newblock In: \emph{2019 International Conference on Robotics and Automation
  (ICRA)}. IEEE, pp. 4303--4309.

\bibitem[{Yu and Tapus(2020)}]{yu2020srg}
Yu C and Tapus A (2020) Srg 3: Speech-driven robot gesture generation with gan.
\newblock In: \emph{2020 16th International Conference on Control, Automation,
  Robotics and Vision (ICARCV)}. IEEE, pp. 759--766.

\bibitem[{Zaraki et~al.(2018)Zaraki, Wood, Robins and
  Dautenhahn}]{zaraki2018development}
Zaraki A, Wood L, Robins B and Dautenhahn K (2018) Development of a
  semi-autonomous robotic system to assist children with autism in developing
  visual perspective taking skills.
\newblock In: \emph{2018 27th IEEE International Symposium on Robot and Human
  Interactive Communication (RO-MAN)}. IEEE, pp. 969--976.

\end{thebibliography}

\end{document}